\documentclass[conference]{IEEEtran}
\IEEEoverridecommandlockouts

\usepackage{cite}
\usepackage{amsmath,amssymb,amsfonts}
\usepackage{algorithmic}
\usepackage{graphicx}
\usepackage{textcomp}
\usepackage{xcolor}
\usepackage{xspace}
\usepackage{multirow}
\usepackage{booktabs}

\makeatletter
\newcommand{\linebreakand}{%
  \end{@IEEEauthorhalign}
  \hfill\mbox{}\par
  \vspace{0.3em}
  \mbox{}\hfill\begin{@IEEEauthorhalign}
}
\makeatother

\newcommand{\ourmethod}{GATE-ST\xspace}

\def\BibTeX{{\rm B\kern-.05em{\sc i\kern-.025em b}\kern-.08em
    T\kern-.1667em\lower.7ex\hbox{E}\kern-.125emX}}
\begin{document}

\title{GATE-ST: Gene-Aware Text-image Encoder for Spatial Transcriptomics}

\author{\IEEEauthorblockN{Lucas Ni}
\IEEEauthorblockA{
\textit{Ridge High School}\\
Basking Ridge, United States \\
lucas.chengming.ni@gmail.com}
\and
\IEEEauthorblockN{Jian Luo}
\IEEEauthorblockA{
\textit{Stony Brook University}\\
Stony Brook, United States \\
jian.luo@stonybrook.edu}

\linebreakand

\IEEEauthorblockN{Wentao Huang}
\IEEEauthorblockA{
\textit{Stony Brook University}\\
Stony Brook, United States \\
wenthuang@cs.stonybrook.edu}
\and
\IEEEauthorblockN{Chao Chen}
\IEEEauthorblockA{
\textit{Stony Brook University}\\
Stony Brook, United States \\
chao.chen.1@stonybrook.edu}
}

\maketitle
\begin{abstract}
Spatial transcriptomics enables spatially resolved gene expression analysis from slide-level images while preserving morphological features, providing valuable information for studying disease mechanisms and developing treatments. However, spatial gene expression profiling typically requires expensive and time-consuming tests. While existing image-based prediction optimizations mostly revolve around including positional embeddings and further image-based changes, text-based optimizations remain relatively unexplored. We present \ourmethod, which incorporates text-based inputs into image-based spatial gene expression predictions. With this approach, generated text descriptions of genes are utilized to better spatial transcriptomics prediction results. Gene summaries are put through a text encoder, generating embeddings that integrate with image embeddings through cross-attention layers to align with morphological features. We demonstrate the effectiveness of such text inputs by benchmarking performance against random gene embeddings and multiple other image-text fusion architectures, and show that \ourmethod outperforms these alternatives. Our results demonstrate the effectiveness of \ourmethod in pathology imaging, which may greatly reduce the time and cost of accurate spatial transcriptomic predictions, proving the potential of text-guided spatial gene expression prediction.
\end{abstract}


\vspace{-3mm}
\section{Introduction}
Spatial gene-expression patterns often correspond to morphological features, tissue composition, and associations with disease. Bulk and single-cell RNA sequencing are currently the leading methodologies for measuring gene expression; however, these methods are costly and compromise morphological feature information. Spatial transcriptomics instead measures gene expression at specific spatial tissue locations while retaining spatial context, allowing it to create associations with tissue morphology. Compared to expensive tests, preparation for spatial transcriptomics only requires widely available and inexpensive hematoxylin and eosin (H\&E)-stained tissue slides. Further development in this field would allow for cheap detection of diseases such as Alzheimer's and cancer, giving great motivation to optimize the ability for spatial transcriptomics to accurately predict gene expression. 

Existing spatial gene-expression prediction methods have largely focused on improving visual representations or incorporating spatial context from morphological features. Although such optimizations have improved performance, other forms of optimization remain unutilized. Further morphological information is limited by the information a histology patch can contain and requires much larger datasets. Comparatively, textual information can provide knowledge about individual genes and their biological functions, contributing entirely new information. Recent text-to-image matching models have surfaced, but comparatively little work has explored this. Incorporating gene descriptions may therefore provide a complementary source of information that can improve existing image-based models.

In this paper, we introduce \ourmethod, Gene-Aware Text-image Encoder for Spatial Transcriptomics, which incorporates gene-text summaries into image-based spatial gene-expression predictions. Given an H\&E patch, we extract morphological features with an image encoder, while encoding textual summaries of target genes. These representations are passed through cross-attention and MLP layers to align the two. This lets text embeddings align with relevant morphological features for model tuning. Our model shows improvements over non text-based models, supporting the potential of text-based optimization.

\section{Related Work}

Early methods predict spatial gene expression directly from individual histology patches. ST-Net~\cite{he2020integrating} applies a pretrained convolutional network followed by a regression head, while HisToGene~\cite{pang2021leveraging} and Hist2ST~\cite{zeng2022spatial} further model relationships among spatial locations using Transformer- and graph-based architectures. Later methods incorporate broader tissue context through multi-resolution or long-range modeling, including M2OST~\cite{wang2025m2ost}, TRIPLEX~\cite{chung2024accurate}, and MERGE~\cite{ganguly2025merge}. Another line of work uses reference-based prediction: BLEEP~\cite{xie2023spatially} aligns image and expression representations through contrastive learning, whereas EGN~\cite{yang2023exemplar} uses retrieved exemplars to refine expression estimates. RankByGene~\cite{huang2026rankbygene} further strengthens image--gene alignment with a cross-modal ranking-consistency loss that preserves the relative ordering of pairwise similarities across modalities. Despite their effectiveness, these methods generally treat genes as a predefined set of output dimensions and do not explicitly model the semantic information associated with individual genes. Recent studies have explored gene names, functional annotations, and phenotype descriptions as additional semantic information. Most of these methods rely on relatively simple image--text fusion. SGN~\cite{yang2024spatial} and AGP-Net~\cite{yang2025agp} predict expression through similarity matching between image and gene-text representations, while GeneQuery~\cite{xiong2024genequery} initially combines projected image and text features through additive fusion before further processing. DeepSpot-M~\cite{nonchev2026deepspot} adopts a more expressive gene-query formulation over image tokens, although textual information is only one of several biological embedding sources used by the model. 

In contrast, \ourmethod uses gene-description embeddings as semantic queries over patch-level visual tokens and introduces image-residual pathways to explicitly preserve morphology-derived information throughout gene-conditioned fusion.

\section{Methodology}

\begin{figure*}[htbp]
    \centering
    \includegraphics[width=0.6\textwidth]{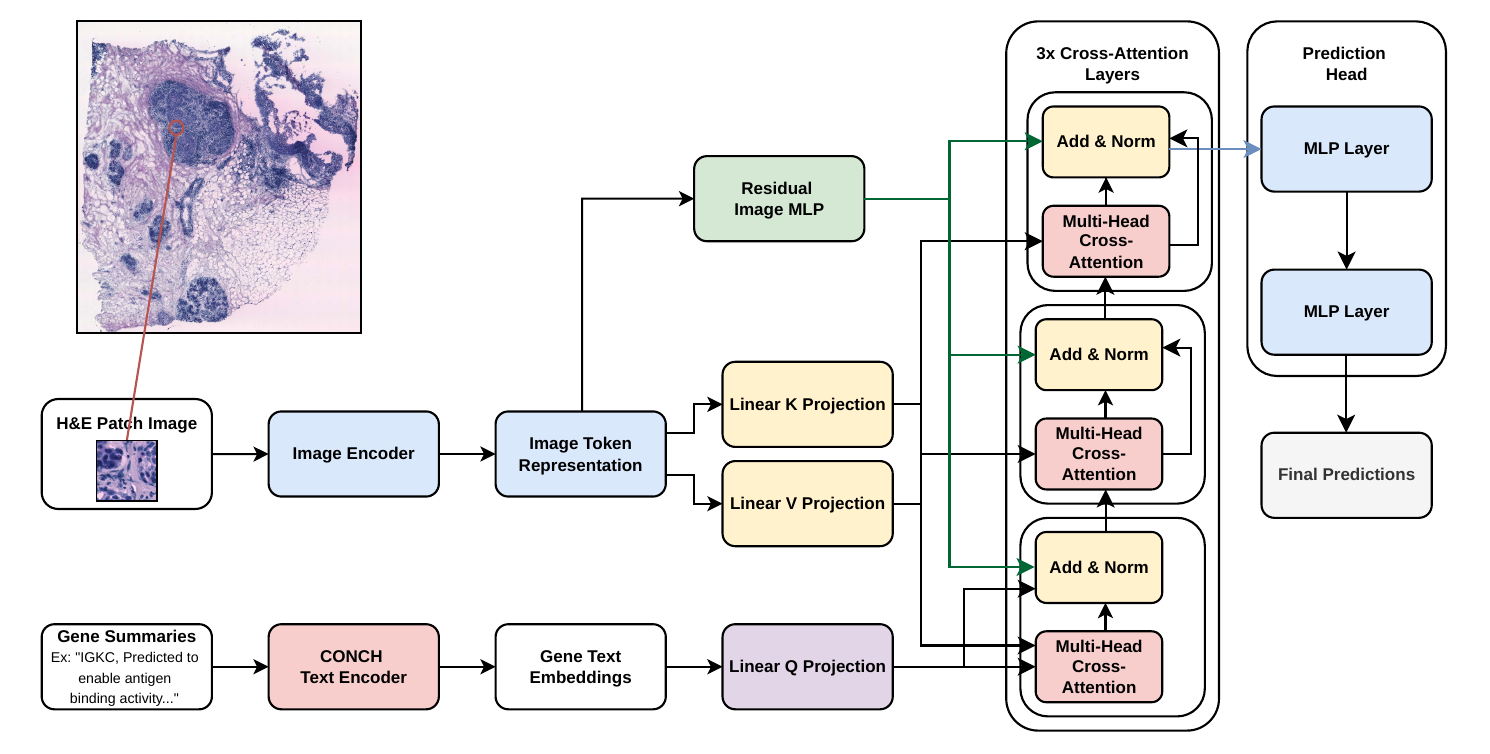}
    \caption{Overview of the model architecture.}
    \label{fig:model_diagram}
\end{figure*}

The proposed method, \ourmethod, predicts spatial gene expression from H\&E-stained images by combining morphological visual information and textual information describing individual genes, as shown in Fig.~\ref{fig:model_diagram}. We use cross-attention layers to align gene descriptions with relevant morphological features, training a distinct representation for each gene. We aim to minimize the difference between our predicted expression and ground-truth measurements. As many genes across the dataset have poor expression, we select the 250 most highly expressed genes as prediction targets, aligning with those in other leading models such as TRIPLEX~\cite{chung2024accurate}. 

An overview of \ourmethod is shown in Fig.~\ref{fig:model_diagram}. The architecture consists of two modules working in parallel prior to the cross-attention architecture: first an image encoder to convert patch images into image embeddings, and then a text encoder to convert gene summaries into text embeddings. The two embeddings are projected into a common feature space and then combined through three cross-attention layers, learning morphological representations for each gene. A global image pathway preserving original patch information also feeds into each cross-attention layer. Afterwards, a final prediction network converts these representations into predicted gene expression values.

\subsection{Image Encoder}
The first module of the architecture, the image encoder, extracts morphological features from the H\&E image patches using a pretrained UNI~\cite{chen2024uni} or CONCH image encoder~\cite{lu2024conch}. Let $B\in\mathbb{N}$ denote the batch size, and $d_i\in\mathbb{N}$ denote the number of pixels along each side of an input image. A batch of image patches is represented as $ I\in\mathbb{R}^{B\times3\times d_i\times d_i}.$

Next, we pass the image features through an image encoder $E_{\mathrm{img}}$. Let $d_s\in\mathbb{N}$ denote the number of spatial image tokens and $d_t\in\mathbb{N}$ denote the dimensionality of each token. For each image $I_i$, the encoder produces image embeddings
$
X_i=E_{\mathrm{img}}(I_i)\in\mathbb{R}^{d_s\times d_t}.
$

For our implementation, UNI inputs are resized to $224\times224$ pixels and CONCH inputs to $448\times448$ pixels using bicubic interpolation. Both start from pretrained checkpoints. UNI consists of 24 updatable Transformer blocks and produces $d_s=196$ spatial tokens with dimensionality $d_t=1024$, while CONCH consists of 12 Transformer blocks and internally produces 784 tokens of dimensionality 768 before pooling them into a single $d_t=512$ image embedding ($d_s=1$). These representations are projected into the common cross-attention dimension $d_c\in\mathbb{N}$. Both encoders may be frozen or updated with LoRA, which we apply to the final 12 blocks of both encoders.

\subsection{Text Encoder}
In parallel, we pass gene-text summaries through a text encoder. Let $d_g\in\mathbb{N}$ denote the number of genes the model makes predictions on. For each gene $g$, we obtain a text summary $s_g$ describing its biological function. Each summary is processed by a pretrained CONCH tokenizer $\tau$, producing representations for each gene. We denote the tokenizer as  $\tau(s_g)\in\mathbb{N}^{L_g},$ where $L_g\in\mathbb{N}$ denotes the length of the tokenized sequence of the summary $s_g$. Next, we pass these tokenized sequences through a pretrained CONCH text encoder $E_{\mathrm{text}}$, producing representations $e_g=E_{\mathrm{text}}(\tau(s_g))\in\mathbb{R}^{d_e}.$
Here, $d_e\in\mathbb{N}$ represents the dimensionality of the text embedding. We define the stacked embeddings of all genes as $G_B$. Prior to the cross-attention layers, we pass each gene embedding through a trainable residual adapter to improve performance. The resulting representations are projected into the common cross-attention dimension $d_c$. As CONCH was pretrained to pair pathology images with textual descriptions, these representations may be trained to align with morphological features generated by the image encoder. 

\subsection{Cross-attention Layer}
\label{sec:cross_attention}
The two representations are now combined in a cross-attention layer, with the image tensor and text embeddings shaped to the common attention dimension $d_c$. We experimentally found that reducing image-embedding dimensionality does not hinder performance. Here, the gene serves as the query and finds relevant morphological features from the spatial image representations, which serve as the keys and values. The corresponding query, key, and value tensors are

\begin{equation*}
\begin{aligned}
Q &= G_BW_Q \in \mathbb{R}^{B\times d_g\times d_c}\\
K &= XW_K \in \mathbb{R}^{B\times d_s\times d_c}\\
V &= XW_V \in \mathbb{R}^{B\times d_s\times d_c}.
\end{aligned}
\end{equation*}

Cross-attention is then computed as

\begin{equation}
A = \operatorname{Softmax}\left( \frac{QK^T}{\sqrt{d_k}} \right).
\end{equation}

Here, $d_k\in\mathbb N$ is the dimensionality of each attention head, and $A\in\mathbb{R}^{B\times d_g\times d_s}$ contains the attention weights relating each gene representation to each image token. Each gene obtains an independent weighting over all of the image representations. These weights are applied to the corresponding vectors, producing $
C=AV\in\mathbb R^{B\times d_g\times d_c}.
$
Thus, $C$ carries an image-conditioned representation for each gene. We use multi-head attention with $h\in\mathbb{N}$ heads and stack $N_c\in\mathbb{N}$ cross-attention layers, passing forward these updated embeddings as the new queries. In our implementation, we use $h=4$ heads, stack $N_c=3$ layers, and a cross-attention dimension $d_c=256$. This combines the benefits of both existing systems with updatable or frozen input layers to align text and image features. Additionally, we add a global image residual pathway to preserve patch information. The image tokens for this are aggregated by mean pooling,

\begin{equation}
\bar X=\frac{1}{d_s}\sum_{j=1}^{d_s}X_{:,j,:}\in\mathbb R^{B\times d_t}.
\end{equation}

The resulting image representation is passed through an image multilayer perceptron, projecting it into the shared cross-attention dimension $d_c$. The resulting feature is broadcast across the $d_g$ gene representations and incorporated at each cross-attention stage. After the final cross-attention layer, we produce $
H^{(N_c)}\in\mathbb R^{B\times d_g\times d_c},
$ where $H$ represents the output of the cross-attention layer. This tensor contains one $d_c$-dimensional representation for each gene. 
\subsection{Prediction Head}
After the final cross-attention layer, each gene is represented by a $d_c$-dimensional feature vector. A shared prediction network $f_\mathrm{pred}$ is applied to each gene independently. We apply this to all genes and image patches to produce a final prediction matrix. For each gene $g$ and image $i$, we have
$
\hat y_{i,g}=f_{\mathrm{pred}}\left(H^{N_c}_{i,g,:}\right)\in\mathbb R.
$ We apply this to all genes to generate a final prediction matrix $\hat Y\in\mathbb R^{B\times d_g}$. In our implementation, the prediction MLP maps each representation to a hidden dimension and then a final scalar output layer. This produces one predicted expression value for each gene and image patch. We use Mean Squared Error (MSE) as the loss function:
$$\mathrm{MSE}_g=\frac{1}{N}\sum_{i=1}^{N}\left(\frac{\hat{y}_{i,g}-\mu_{\hat{y}_g}}{\sigma_{\hat{y}_g}}-\frac{y_{i,g}-\mu_{y_g}}{\sigma_{y_g}}\right)^2.
$$

\section{Experiments}

\textbf{Dataset \& Preprocessing} 
Experiments were conducted on the HER2+ breast cancer dataset~\cite{jaume2024hest}, consisting of 36 spatially profiled tissue samples from 8 patients with HER2-positive breast tumors. In the original dataset, donors were labelled A-H, with four patients providing 6 samples each and the other four providing 3 samples each. These sections were hematoxylin-and-eosin stained at $\times 20$ magnification, providing the images used for training and testing. The dataset includes 13,136 total spots with a diameter of $100 \mu m$. We smooth the expressions with 8-neighborhood smoothing, averaging the gene expression values of each spot with its eight neighbors from a local $3\times 3$ patch collection following MERGE~\cite{ganguly2025merge}. This reduces technical dropouts and sparsity in spatial transcriptomic measurements~\cite{ganguly2025merge}. We analyze the 250 most highly expressed genes following the gene-selection protocol used in TRIPLEX~\cite{chung2024accurate}.

\textbf{Baselines}
We evaluate \ourmethod using both UNI~\cite{chen2024uni} and CONCH~\cite{lu2024conch} as image encoders and compare with five baseline architectures. 1) \textbf{Multi-layer Perceptron (MLP):} this method uses only the image encoder by passing image embeddings through two linear layers to make predictions; 2) \textbf{Direct Feature Concatenation:} this method directly concatenates the text and image embeddings, then passes the concatenated representation through two linear layers to produce final predictions; 3) \textbf{Embedding-Image Similarity Based Predictions:} this method calculates cosine similarity between the image and text embeddings, then passes the similarity scores through an MLP layer to make final predictions; 4) \textbf{Cross-attention Fusion:} We replace direct feature concatenation with cross-attention layers between the two. Summary embeddings serve as query vectors and image embeddings produce key and value vectors, passing through three cross-attention layers and a final prediction MLP layer to make predictions; 5) \textbf{Cross-attention with image-residual MLP:} This model extends the cross-attention fusion baseline to include an image representation fed back at each layer. Image tokens are averaged to produce a global image feature that passes through two MLP layers to add to each cross-attention layer, with the remaining architecture unchanged.

\textbf{Metrics}
We evaluate performance using Mean Squared Error (MSE) and Pearson correlation. For each gene, we standardize the predicted and actual expression values across the spots to scale this metric. Final MSE is the mean of the squared errors for each gene, reducing differences in absolute expression scale between genes. Pearson correlation is calculated between predicted and measured expression across spots for each gene.
Optimizations are aimed at improving these two metrics. Both metrics are averaged across the 250 genes.

\begin{figure}[htbp]
    \centering
    \includegraphics[width=0.4\textwidth,trim=0 3mm 0 9.5mm, clip]{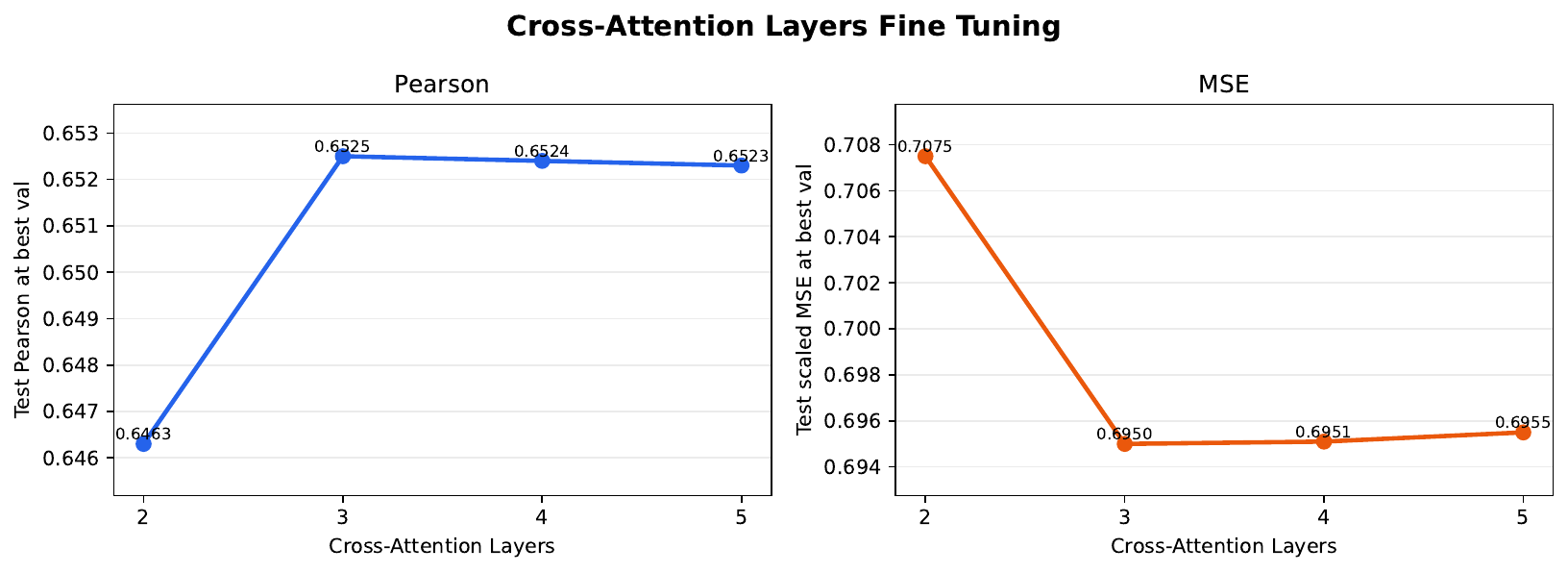}
    \caption{Cross-Attention Layers Fine Tuning}
    \vspace{-3mm}
    \label{fig:ablation_study_layers}
    \vspace{-2mm}
\end{figure}

\textbf{Hyperparameters}
Hyperparameters were selected by grid search. For example, three cross-attention layers consistently outperformed other settings across the various architectures we tested, as shown in Fig.~\ref{fig:ablation_study_layers}. A comprehensive list of our hyperparameters is shown in Table~\ref{tab:hyperparameters}.

\begin{table}[htbp]
    \centering
    \caption{Final hyperparameter configurations for models using UNI and CONCH image encoders.}
    \label{tab:hyperparameters}
    \resizebox{0.25\textwidth}{!}{%
    \begin{tabular}{lcc}
        \toprule
        \textbf{Hyperparameter} & \textbf{UNI} & \textbf{CONCH} \\
        \midrule

        Batch size                  & 32              & 32 \\
        Learning rate               & $3\times10^{-4}$ & $3\times10^{-4}$ \\        Maximum epochs              & 200             & 200 \\
        Prediction-head dropout     & 0.4             & 0.4 \\

        LoRA rank ($r$)             & 8               & 8 \\
        LoRA scaling ($\alpha$)     & 16              & 16 \\

        \bottomrule
    \end{tabular}%
    }
    \vspace{-1mm}
\end{table}

\subsection{Main Experiment}
For our experiments we used 8-fold donor-level cross-validation. Model selection was based on the highest-performing model on the validation split. For each gene, the measured and predicted values are standardized prior to metric calculation. Results of our experiments are shown in Table~\ref{tab:conch_uni_results}. For both the CONCH and UNI image encoders, \ourmethod outperforms all other text-integration methods. \ourmethod yields a 0.7280 MSE and 0.6360 Pearson coefficient with the CONCH image encoder, and 0.6774 MSE and 0.6613 Pearson with the UNI image encoder. Compared with direct feature concatenation, \ourmethod yields a 0.1006 lower MSE and 0.0503 higher Pearson coefficient with the CONCH image encoder, and a 0.0194 lower MSE and 0.0097 higher Pearson coefficient, supporting its effectiveness.

Text encoding likely carries useful additional information that better gene predictions. As \ourmethod contains cross-attention, such features can be learned to match with morphological features, providing more information and improving predictions. Further information added by textual inputs is shown to improve results, as each of our optimizations that allows for more incorporation of textual inputs improves results. However, stronger image encoders may benefit less from such additions, as shown with the smaller improvements when using the UNI image encoder. 
\begin{table}[t]
    \centering
    \caption{Comparison of gene-expression prediction architectures using
    CONCH and UNI image encoders.}
    \label{tab:conch_uni_results}

    \resizebox{0.45\textwidth}{!}{%
    \begin{tabular}{lcccc}
        \toprule
        \multirow{2}{*}{\textbf{Methods}}
        & \multicolumn{2}{c}{\textbf{CONCH}~\cite{lu2024conch}}
        & \multicolumn{2}{c}{\textbf{UNI~\cite{chen2024uni}}} \\

        \cmidrule(lr){2-3}
        \cmidrule(lr){4-5}

        & \textbf{MSE} $\downarrow$
        & \textbf{Pearson} $\uparrow$
        & \textbf{MSE} $\downarrow$
        & \textbf{Pearson} $\uparrow$ \\

        \midrule

        Pure MLP
        & 0.8180 & 0.5910 & \textbf{0.6731} & \textbf{0.6635} \\

        Direct Feature Concatenation
        & 0.8286 & 0.5857 & 0.6968 & 0.6516 \\

        Embedding-Image Similarity Based Predictions
        & 0.8217 & 0.5892 & 0.6822 & 0.6589 \\
    
        Cross-attention Fusion
        & 0.8009 & 0.5996 & 0.6798 & 0.6601 \\

        Cross-attention with image-residual MLP
        & \underline{0.7867}
        & \underline{0.6066}
        & 0.6785
        & 0.6608 \\

        \textbf{\ourmethod(Our)}
        & \textbf{0.7280}
        & \textbf{0.6360}
        & \underline{0.6774}
        & \underline{0.6613} \\

        \bottomrule
    \end{tabular}%
    }
    \vspace{-1mm}
\end{table}

\begin{figure}[htbp]
    \centering
    \includegraphics[width=0.4\textwidth]{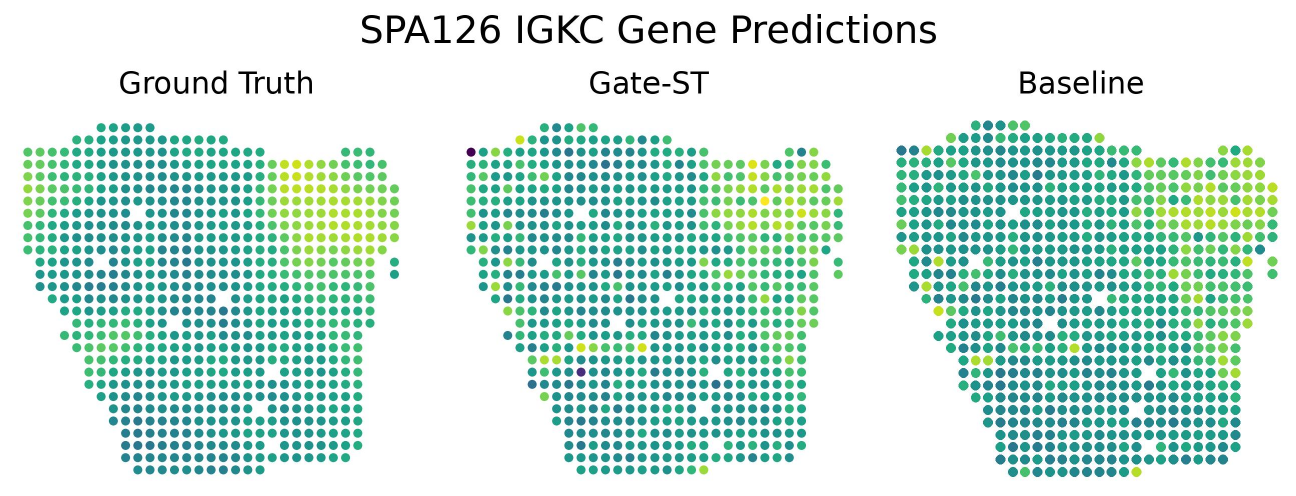}
    \caption{Comparison Heatmaps of the IGKC Gene Expression.}
    \label{fig:heatmaps}
\end{figure}

\textbf{Qualitative Results}
We plot gene expression for specific relevant genes to compare our predictions with the ground truth as shown in Fig.~\ref{fig:heatmaps}. These heatmaps show the gene expression of the IGKC gene across a particular slide in the dataset. It is clear that the predictions show strong correspondence with the ground truth, supporting the effectiveness of \ourmethod.

\vspace{-2mm}

\subsection{Ablation}

Prior experiments outlined in Table~\ref{tab:conch_uni_results} provide quantitative comparisons and contributions of model changes, but we also quantitatively analyze the results by freezing updatable portions of the encoder. We first test whether improvements come from model architecture changes or if textual inputs carry true information, so for this we replace our CONCH text embeddings with random Gaussian-distributed orthogonal vectors that carry no information, where improved performance over this baseline would imply text embeddings were helpful. Otherwise, this follows the same architecture as the fully optimized model. This also indicates that this method of changing the architecture with random embeddings does not improve results, attributing the improvement to textual input. Table~\ref{tab:ablation_components} compares this ablation with the complete model.
\vspace{-4mm}
\begin{table}[htbp]
    \centering
    \caption{Ablation study of individual components in \ourmethod using CONCH and UNI image encoders.}
    \label{tab:ablation_components}
    \resizebox{0.48\textwidth}{!}{%
    \begin{tabular}{lccc|cc|cc}
        \toprule
        \multirow{2}{*}{\textbf{Model Variant}}
        & \multirow{2}{*}{\textbf{Text Embeddings}}
        & \multirow{2}{*}{\textbf{Image Residual}}
        & \multirow{2}{*}{\textbf{LoRA}}
        & \multicolumn{2}{c|}{\textbf{CONCH}}
        & \multicolumn{2}{c}{\textbf{UNI}} \\

        \cmidrule(lr){5-6}
        \cmidrule(lr){7-8}

        & & &
        & \textbf{MSE} $\downarrow$
        & \textbf{Pearson} $\uparrow$
        & \textbf{MSE} $\downarrow$
        & \textbf{Pearson} $\uparrow$ \\

        \midrule

        Random Gene Embeddings
        & -- & \checkmark & \checkmark
        & 0.8191 & 0.5905
        & 0.7338 & 0.6311 \\

        Cross-Attention
        & \checkmark & -- & --
        & 0.8009 & 0.5996
        & 0.6798 & 0.6601 \\

        + Image Residual
        & \checkmark & \checkmark & --
        & 0.7867 & 0.6066
        & 0.6785 & 0.6608 \\

        \textbf{Full Model}
        & \checkmark & \checkmark & \checkmark
        & \textbf{0.7280} & \textbf{0.6360}
        & \textbf{0.6774} & \textbf{0.6613} \\

        \bottomrule
    \end{tabular}%
    }
    \vspace{-2mm}
\end{table}

As shown in Table~\ref{tab:ablation_components}, the ablation study with informationless vectors performs worse than standard experiments with true text embeddings, with the CONCH system reporting an MSE of 0.8191 and Pearson correlation of 0.5910 with the Gaussian distribution, and an MSE of 0.7280 and Pearson of 0.6360 with our final optimizations. The UNI baseline was at 0.7338 MSE and 0.6311 Pearson with the Gaussian distribution, but 0.6774 MSE and 0.6613 Pearson with our optimizations. There exists a significant improvement between these two models, which can be attributed to textual inputs. This means textual inputs carry meaningful information for morphological features to align with, which suggests opportunities for future work. This table further proves that the full model also improves with cross-attention, LoRA adaptation, and an added global image residual, justifying such changes to our architecture.

\subsection{Hyperparameter Tuning}
\begin{figure}[htbp]
    \centering
    \includegraphics[width=0.4\textwidth,trim=0 3mm 0 9.5mm, clip]{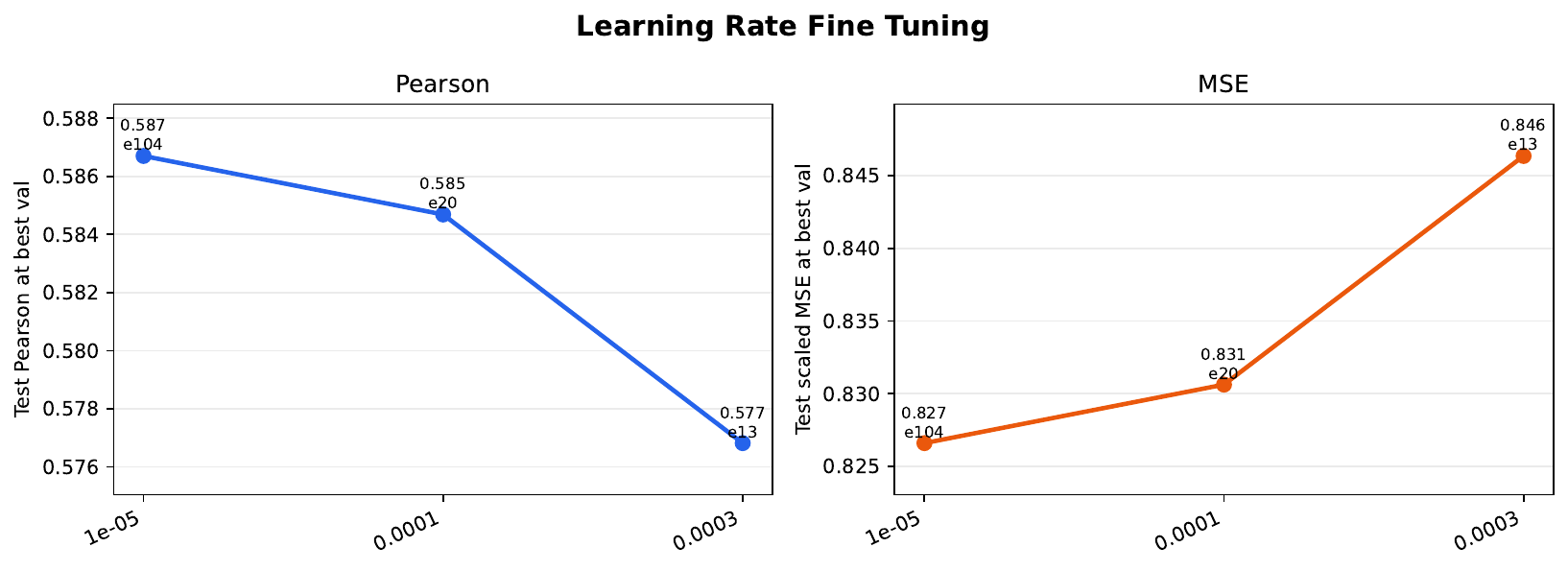}
    \caption{Learning Rate Fine Tuning}
    \label{fig:learning_rates}
    \vspace{-3mm}
\end{figure}

Various hyperparameters such as learning rates and the number of cross-attention layers were tested for model stability across hyperparameter changes. Relatively stable performance across different numbers of layers and hyperparameters proves its stability and robustness to hyperparameter variation. As shown in Fig.~\ref{fig:learning_rates}, varying the number of cross-attention layers after two and changing learning rates have little to no effect on the model performance. Across the tested settings, this shows the model is resistant to small hyperparameter adjustments. Our final configurations were chosen based on the best-performing settings, but our experiments show such changes do not meaningfully impact the results. 

\vspace{-2mm}
\vspace{-2mm}

\section{Conclusion}
In this paper, we present \ourmethod, a new approach to spatial transcriptomics utilizing text-based gene inputs which remain relatively unexplored. Previous spatial transcriptomics optimization methods focus on positional encoding or purely image or morphology-based optimizations. However, few studies have explored text-based optimization for spatial transcriptomics. We build on such models and present improvements over non-text inputs once they are included for querying. Text-based inputs improve performance and outperform random embeddings, proving their information-carrying nature. Such descriptions will be readily and inexpensively available to medical professionals, offering very great potential for future developments on this type of optimization. Despite such benefits, opportunities remain for improving text implementation and evaluating the approach on larger datasets, as our experiments were conducted on a relatively small dataset. However, our paper has proved the potential of text-based inputs in optimizing spatial transcriptomics predictions.
\vspace{-1mm}
\bibliography{IEEEexample}
\bibliographystyle{IEEEtran}

\end{document}